\documentclass{article}

     \PassOptionsToPackage{numbers, compress}{natbib}

    \usepackage[preprint]{neurips_2024}

\usepackage[utf8]{inputenc} 
\usepackage[T1]{fontenc}    
\usepackage{hyperref}       
\usepackage{url}            
\usepackage{booktabs}       
\usepackage{amsfonts}       
\usepackage{nicefrac}       
\usepackage{microtype}      
\usepackage{xcolor}         
\usepackage{booktabs}
\usepackage{multirow}

\usepackage{amssymb}
\usepackage{amsmath}
\usepackage{caption}
\usepackage{subcaption}
\usepackage{amsmath}
\usepackage{graphicx}
\usepackage{algorithm2e}
\usepackage{algorithmic}

\usepackage[capitalize]{cleveref}
\crefname{section}{Sec.}{Secs.}
\Crefname{section}{Section}{Sections}
\Crefname{table}{Table}{Tables}
\crefname{table}{Tab.}{Tabs.}

\title{A Generative Framework for Self-Supervised Facial Representation Learning}

\author{%
  Ruian He, Zhen Xing, Weimin Tan\footnote{Corresponding authors: Bo Yan, Weimin Tan.}, Bo Yan\footnotemark[1] \\
  School of Computer Science, Shanghai Key Laboratory of Intelligent Information Processing, \\
  Fudan University, Shanghai \\
  \texttt{\{rahe16, xingz20, wmtan, byan\}@fudan.edu.cn} \\
}

\begin{document}

\maketitle

\begin{abstract}
Self-supervised representation learning has gained increasing attention for strong generalization ability without relying on paired datasets. However, it has not been explored sufficiently for facial representation. Self-supervised facial representation learning remains unsolved due to the coupling of facial identities, expressions, and external factors like pose and light. Prior methods primarily focus on contrastive learning and pixel-level consistency, leading to limited interpretability and suboptimal performance. In this paper, we propose LatentFace, a novel generative framework for self-supervised facial representations. We suggest that the disentangling problem can be also formulated as generative objectives in space and time, and propose the solution using a 3D-aware latent diffusion model. First, we introduce a 3D-aware autoencoder to encode face images into 3D latent embeddings. Second, we propose a novel representation diffusion model to disentangle 3D latent into facial identity and expression. Consequently, our method achieves state-of-the-art performance in facial expression recognition (FER) and face verification among self-supervised facial representation learning models. Our model achieves a 3.75\% advantage in FER accuracy on RAF-DB and 3.35\% on AffectNet compared to SOTA methods. 
\end{abstract}

\section{Introduction}
\label{sec:intro}

Human faces hold significant importance in the field of computer vision, serving as carriers of vital information, including identity, expressions, and intentions  \cite{hasselmo1989role}. Recent advancements in neural networks have led to remarkable achievements in facial understanding  \cite{Cao2018VGGFace2AD,Wang2020RegionAN}. Nonetheless, supervised learning requires large-scale annotated datasets and faces challenges such as imbalance annotation  \cite{zeng2018facial,gu2022tackling} and the labor-intensive labeling work  \cite{kollias2022abaw}. In response, self-supervised methods  \cite{Wiles2018SelfsupervisedLO} has been proposed to tackle this issue by focusing on learning representations rather than directly predicting labels.

General image representation learning methods, such as contrastive learning  \cite{chen2020simple} and masked autoencoders \cite{he2022masked}, have garnered considerable interest. These models are trained on unlabeled datasets to acquire implicit representations. Nevertheless, the general representations lack interpretability and fail to account for the semantic and structural aspects of the human face.

\begin{figure}[t]
	\centering
    \includegraphics[width=\linewidth]{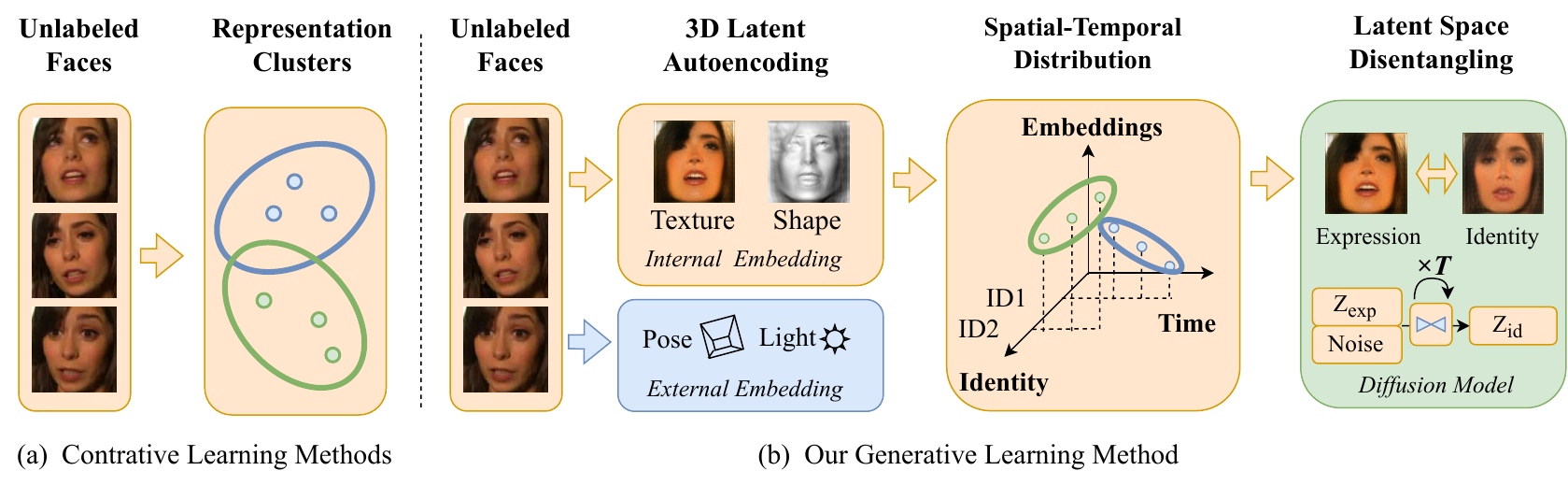}
	\caption{Comparison of different learning paradigm. Our generative framework enables better spatial-temporal awareness and more thorough representation disentangling than previous contrastive learning methods. }
    \label{fig:teaser}
\end{figure}

To leverage the face structure, disentangling is a common practice for learning facial representations. The approach is to distinguish facial expression and identity from changing environments. Prior methods  \cite{Li2019SelfSupervisedRL,Lu2020SelfSupervisedLF,Chang_2021_ICCV,Shu2022RevisitingSC} have achieved good results using contrastive learning. However, there are two problems with contrastive facial representation disentangling. First, previous techniques exhibit limitations in terms of spatial awareness and have subpar performance when there is drastic light and pose variations of face images. Second, they focus on pixel-level warping  \cite{Li2019SelfSupervisedRL,Chang_2021_ICCV}, leading to the incomplete disentanglement of facial expression. To solve the problems, we suggest that the disentangling problem can be also formulated as generative objectives in space and time.

Recent advances in generative models, especially those based on diffusion models  \cite{Ho2020DenoisingDP,SohlDickstein2015DeepUL}, boost image restoration  \cite{Kawar2022DenoisingDR}, semantic segmentation \cite{Xu2023OpenVocabularyPS} and structure prediction \cite{abramson2024accurate}, which reveal the novel potential of representation learning. Inspired by this, we explore generative approaches in self-supervised facial representation learning. Our contributions to address the problems of previous works are three-fold:

First, we propose LatentFace, a novel generative framework for self-supervised facial representations, as illustrated in \cref{fig:teaser}. To the best of our knowledge, it is the first 3D-aware latent diffusion model for self-supervised facial representation learning, which has better disentangling ability.

Second, our model incorporates 3D Latent Autoencoding and Latent Space Disentangling. The 3D Latent Autoencoding disentangles the face latent codes in space, while the Latent Space Disentangling predicts the facial identity as the time-invariant part of face latent codes.

Third, extensive experiments have demonstrated that our model achieves state-of-the-art performance on unsupervised facial expression recognition (FER) and face verification. Specifically, our model has a 3.75\% advantage in FER accuracy on RAF-DB  \cite{li2017reliable} and 3.35\% on AffectNet  \cite{Mollahosseini2019AffectNetAD} compared to SOTA unsupervised methods.

\section{Methodology}
\subsection{Revisiting Facial Representation Learning}
\label{sec:framework}

As illustrated in \cref{fig:teaser}, our proposed generative framework learns facial expression and identity embeddings from unlabeled images and videos. Given an image $I$, the facial representation learning models are expected to encode the image as facial embeddings. The facial embedding should be first disentangled from the environment, such as light and facial pose, which can be modeled with a 3D auto-encoding model \cite{Wu2020UnsupervisedLO}:
\begin{equation}
    \min _{\theta,\phi} \mathbb{E}_I\left[d\left(x, D_{\theta}\left(E_{\phi}(x)\right)\right)\right]
\end{equation}
where $I$ is the image data, $D, E$ are the decoders, and the encoders of pose, light, and facial embeddings. $\theta$ and $\phi$ are the model parameters and $d$ is the distance between the reconstructed image and the original image, usually including photometric loss and perception loss. Next, we consider the discrimination of the facial identity $Z_{id}$ and the facial expression $Z_{exp}$. In traditional parametrized face models \cite{Blanz1999AMM}, the facial expression is modeled as the deviation of an emotional face from the facial identity. Specifically, the parameters $Z_{exp}$ are for faces of facial expression, and the parameters $Z_{id}$ are for faces with neutral expression, which also refers to facial identity. And the facial expression $\Delta_{exp}$ is taken as a bias of neutral face and emotional face:
\begin{equation}
Z_{exp} = Z_{id} + \Delta_{exp}
\end{equation}
Therefore, the challenge of disentangling facial identity and expression lies in predicting the identity embedding with a given image, which can be learned from the temporal information of videos.

We consider the identity embedding $Z_{id}$ should be the same in different frames $I_1, I_2$ of a video of the same person. Therefore, the optimal embedding $Z_{id}^*$ should meet the following formula.
\begin{equation}
Z_{id}^* = \underset{Z_{id}}{\operatorname{argmax}}\ {p(Z_{id} \mid I_1, I_2)} = \underset{Z_{id}}{\operatorname{argmax}}\{\underbrace{\log p(I_1, I_2 \mid Z_{id})}_{\text {data term}} + \underbrace{\log p\left(Z_{id}\right)}_{\text {generation prior}} \}
\end{equation}
where $\log p(I_1, I_2 \mid Z_{id})$ depends on the construction of input data, while $\log p\left(Z_{id}\right)$ is the generation prior of $Z_{id}$. Therefore, how to extract the facial embedding and how to generate $Z_{id}$ becomes the problem. We separately address them with 3D Latent Autoencoding and Latent Space Disentangling.

\subsection{3D Latent Autoencoding} 
\label{sec:pixel}

\begin{figure*}[t]
\centering
\includegraphics[width=\linewidth]{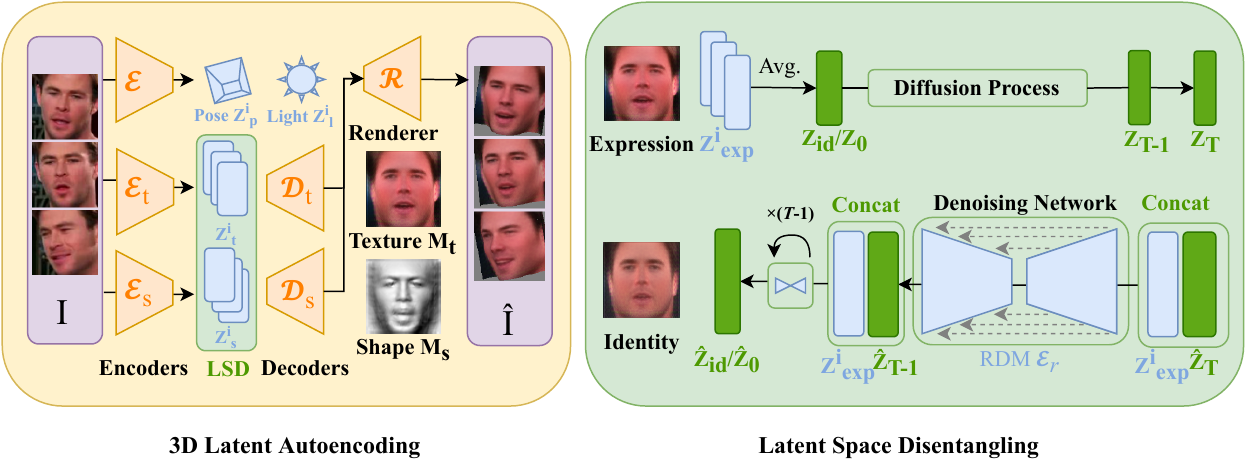}
\caption{Overview of the proposed framework. In the first stage, we disentangle 3D factors, including texture, shape, pose, and light, through the training of autoencoders (comprising encoders $\mathcal{E}$ and decoders $\mathcal{D}$) and render them using a renderer $\mathcal{R}$ to reconstruct the input. In the second stage, we further disentangle the texture and shape latent and train the Representation Diffusion Model (RDM) $\mathcal{E}{r}$ to generate the identity latent $Z_{id}$ from the emotional face latent $Z_{exp}$.}
\label{fig:framework}
\end{figure*}

As shown in \cref{fig:framework}, our framework comprises two stages: 3D latent autoencoding and latent space disentangling. The first stage of our approach is to disentangle 3D facial embeddings, like facial texture $Z_t$ and shape $Z_s$, from environmental embeddings, including pose $Z_p$ and light $Z_l$. Leveraging the ample variations in facial and external factors found in image sets  \cite{liu2015faceattributes}, the model learns to disentangle them with an autoencoder. We follow Unsup3D  \cite{Wu2020UnsupervisedLO} to model the face as an unsupervised non-linear parametric model of texture and shape. The facial representations are learned through a self-supervised scheme, which uses the minimal assumption of face symmetry to make the problem converge to stereo reconstruction. 

Specifically, the texture and shape of 3D face model are learned by encoders $\mathcal{E}$ and decoder $\mathcal{D}$ with the equation $M = \mathcal{D}(Z) = \mathcal{D}(\mathcal{E}(I))$. In contrast, the light and pose are predicted directly from encoders with the equation $Z=\mathcal{E}(I)$. Finally, the model is trained to reconstruct the input face $I$ by rendering the extracted features and the flipped feature map:
\begin{align}
\label{eqa:rendering}
\hat{I} &= \mathcal{R}(M_t,M_s,Z_p,Z_l),  \\
\hat{I}' &= \mathcal{R}(\text{flip} (M_t),\text{flip} (M_s),Z_p,Z_l),
\end{align}
where $M_s$ is the facial shape and $M_t$ is the facial texture.  $Z_l$ is the light color and direction, and $Z_p$ is the camera's pose. $\text{flip}(\cdot)$ is a horizontal flip operator. $\hat{I}$ is the reconstructed face, and $\hat{I}'$ is the symmetric face generated from the flipped feature map. For the renderer $\mathcal{R}$, we implement a differentiable rendering pipeline with a diffuse illumination model and a weak perspective camera using Pytorch3D \cite{ravi2020pytorch3d}. We provide detailed implementation in the supplementary materials.

\subsection{Latent Space Disentangling}
\label{sec:latent}

Beyond autoencoding for 3D factors, we further train the model to disentangle facial expressions and identity. While previous work  \cite{Li2019SelfSupervisedRL,Chang_2021_ICCV} models the facial expression as the pixel-level warping, we propose to predict the time-invariant mean in the latent space, as depicted in \cref{fig:teaser}. 

In the second stage, we propose to leverage unlabeled video sequences to construct the learnable data term $\log p(I_1, I_2 \mid Z_{id}
)$. Notably, we observe that the web face videos  \cite{Nagrani2017VoxCelebAL} often feature individuals of the same identity with varying facial expressions. Following the neutral face assumption \cite{Blanz1999AMM}, we approximate the identity $Z_{id}$ with the average latent embedding of $Z_{exp}$ in a video sequence, which can be taken as a time-invariant part of the facial attribute. 

Specifically, to take texture embedding $Z_t$ as an example, we suppose the autoencoder ($\mathcal{E}_t$ and $\mathcal{D}_t$) has learned a complete representation of facial texture in the image set. And for every frame $I_i, i = 1, ..., n$ and $n$ is the number of sampled frames, we can get the encoded embedding $Z_t^i$ for a compressed representation of facial texture. We collect the frames sparsely in a video so that the expression changes significantly to discriminate, and suppose the distribution of $\overline{Z_t}$ can approximate that of $Z_{id}$.  Given the encoded embedding $Z_t^i$ from a video sequence, we denote the embedding $Z_t^i$ as the emotional face embedding $Z_{exp}$ and the average embedding $\overline{Z_t}$ of the sampled sequence as the target facial identity $Z_{id}$.

	

\subsection{Representation Diffusion Model}

Inspired by recent success of latent diffusion models  \cite{Rombach2021HighResolutionIS}, we propose a Representation Diffusion Model (RDM) to generate the facial identity latent with the facial shape and texture latent as the condition.  \cref{fig:framework} shows the pipeline of RDM. 

For the diffusion process, we sample a noisy embedding $Z_\tau$ at time step $\tau$ from the identity embedding $Z_0=Z_{id}$ as:
\begin{equation}
    Z_\tau = \sqrt{\bar{\alpha}_{\tau}} Z_0+\sqrt{1-\bar{\alpha}_{\tau}} \epsilon, \quad \epsilon \sim \mathcal{N}(0, 1)
\end{equation}
where $\tau$ is the diffusion step we use, $\alpha_1, ..., \alpha_T$ is a predefined noise schedule and $\bar{\alpha}_\tau=\prod_{k=1}^\tau\alpha_k$ as defined in  \cite{ho2020denoising}. $\epsilon$ is the artificial noise sample from a Gaussian distribution $N(0,1)$. As the training examples, we randomly add noise to the latent to a diffusion step $\tau$ ranging from 1 to $T=1000$, \textit{i.e.}, $\tau \sim \text { Uniform }(1, T)$. 

In the training time, a denoising network $\mathcal{E}_{r}$ is trained to remove the artificial noise $\epsilon$ with the original latent $Z_{exp}$ as the condition. We concatenate the $Z_{exp}$ with the noisy latent $Z_t$ and feed them into $\mathcal{E}_{r}$ to predict the clean latent $\hat{Z}_0= \mathcal{E}_{r}\left(Z_\tau, \tau, Z_{exp}\right)$. The optimization objective of the UNet can be expressed as follows:
\begin{equation}
\label{eqa:unet}
    \min _{\theta} \mathbb{E}_{Z_{0}, \varepsilon \sim N(0, 1), \tau \sim \text { Uniform }(1, T)}\left\|Z_0-\mathcal{E}_{r}\left(Z_\tau, \tau, Z_{exp}\right)\right\|^{2},
\end{equation}
where $Z_{exp}$ is the embedding of the emotional face and $Z_\tau$ is a noisy sample of $Z_0$ at timestep $\tau$. $\theta$ is the parameters of $\mathcal{E}_{r}$. RDM is also trained for facial shape embedding $Z_s$. In the inference time, we use the DDIM \cite{Song2020DenoisingDI} sampler at $S=5$ steps for generating the identity latent $\hat{Z}_0$ from a random noise $\hat{z}_T \sim N(0, 1)$. Then we can get the disentangled expression latent $\Delta_{exp}$ by subtracting $\hat{Z_{0}}$ from $Z_{exp}$.

\begin{figure}[t]
  \centering
  \includegraphics[width=0.8\linewidth]{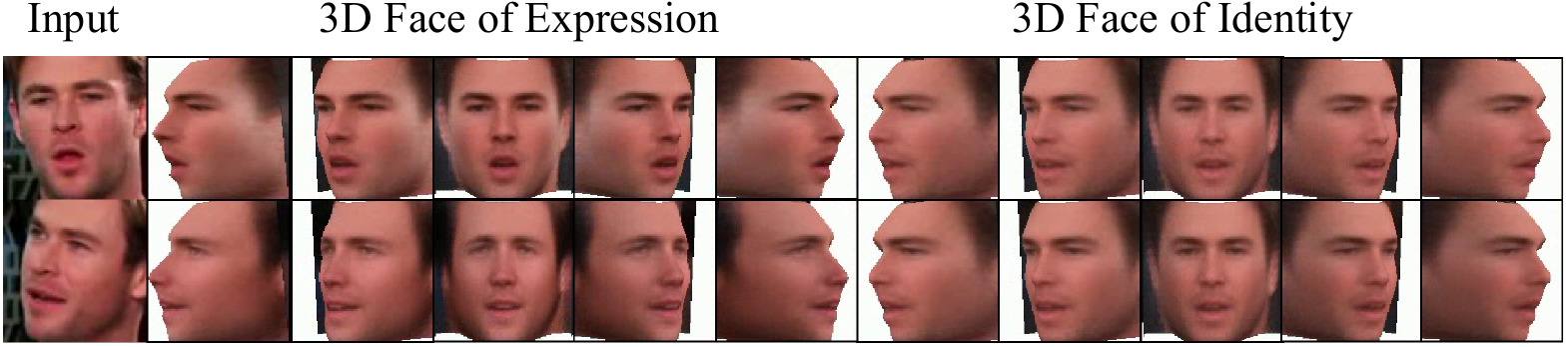}
   \caption{Visualization of generated facial identity. We show the reconstructed 3D face of 2 frames in the input video sequence. The facial identity is disentangled from expressions.}
   \label{fig:first}
\end{figure}

\section{Experiments}
\label{sec:comparison}

\subsection{Implementation Details} 

\paragraph{Architecture} Our proposed model is implemented based on the PyTorch framework. The encoders and decoders are standard fully-convoluted networks with batch normalization layers. The embedding size of $Z_s$ and $Z_t$ are set as 256. The denoising UNet \cite{Ronneberger2015UNetCN} is designed as 3 blocks with two ResNet \cite{He2016DeepRL} layers in each block.  Detailed network implementation and loss functions can be found in the supplementary material. Training and inferencing codes are also provided.

\paragraph{Training Datasets} Our model is trained on CelebA  \cite{liu2015faceattributes} dataset for the first stage and Voxceleb  \cite{Nagrani2017VoxCelebAL} for the second stage. CelebA has 10,177 identities and 202,599 face images. We follow the official split to use 162,257 images for training. VoxCeleb has in total 153,516 video clips of 1,251 speakers. We select 1147 speakers as the training set following this work \cite{Nagrani2018SeeingVA}. We crop the images in CelebA and Voxceleb datasets with MTCNN  \cite{Schroff2015FaceNetAU} and resize them to 64 $\times$ 64. 

\paragraph{Training Procedure} The model is trained with the Adam optimizer  \cite{Kingma2015AdamAM}. The model is trained for 30 epochs, in the first stage and the second stage respectively. We set the batch size to 16, and the learning rate is 0.0001 for both training stages. In the second stage, we randomly sample 16 frames in a video sequence for training. The experiments are performed on a server with RTX 3090 GPUs.

\subsection{Evalutation Settings}

\paragraph{Baselines} We compare our model with state-of-the-art self-supervised methods, FAb-Net  \cite{Wiles2018SelfsupervisedLO}, TCAE  \cite{Li2019SelfSupervisedRL}, Temporal  \cite{Lu2020SelfSupervisedLF}, FaceCycle  \cite{Chang_2021_ICCV}, SSLFER  \cite{Shu2022RevisitingSC} and PCL \cite{liu2023pose}. We adopt the officially released models by the authors. Moreover, we also leverage state-of-the-art self-supervised image representation learning methods, the constrative learning method SimCLRv2  \cite{chen2020simple} and the masked autoencoders  \cite{he2022masked}, as baselines. We use ResNet50  \cite{He2016DeepRL} model for SimCLR and ViT-Base  \cite{dosovitskiy2020image} model for MAE. The models are trained on CelebA and VoxCeleb dataset separately for 30 epochs at $224\times 224$ resolution. Finally, we compare with the state-of-art unsupervised face model Unsup3D  \cite{Wu2020UnsupervisedLO} trained on CelebA to show the effectiveness of latent disentanglement.

\paragraph{Evaluation Protocol} We adopt a commonly-used linear probing protocol  \cite{he2022masked} for representation evaluation. The input is first resized to the corresponding resolution of each baseline. We freeze the backbone feature extraction network and train a batch normalization layer followed by a linear layer after the extracted features to reduce the feature dimension and match the output. The linear layer is trained with Adam \cite{Kingma2015AdamAM} optimizer and a learning rate of 1e-3.

\begin{figure}[t]
	\centering
	\includegraphics[width=0.85\linewidth]{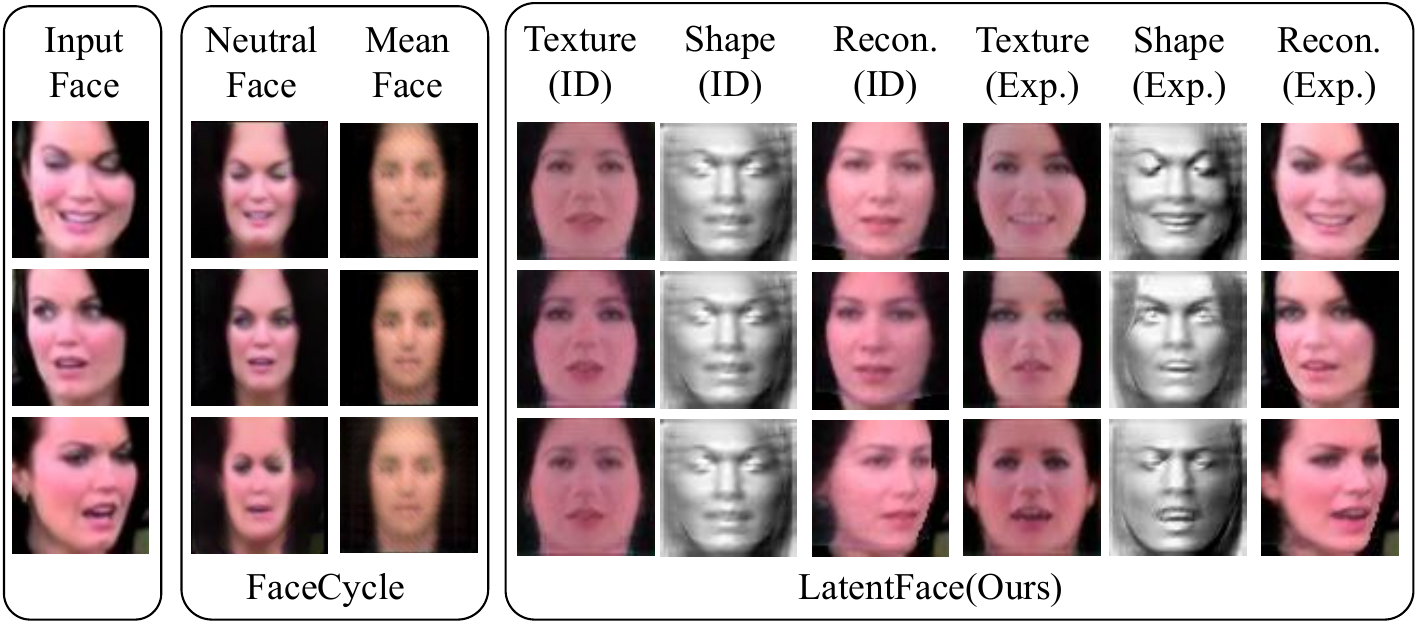}
	
	\caption{Comparison of disentangled representations. Our method have a more detailed and fidelity representation than the SOTA method FaceCycle \cite{Chang_2021_ICCV}.}
	\label{fig:disentangle}
\end{figure}

\begin{figure}[t]
	\centering
	\includegraphics[width=\linewidth]{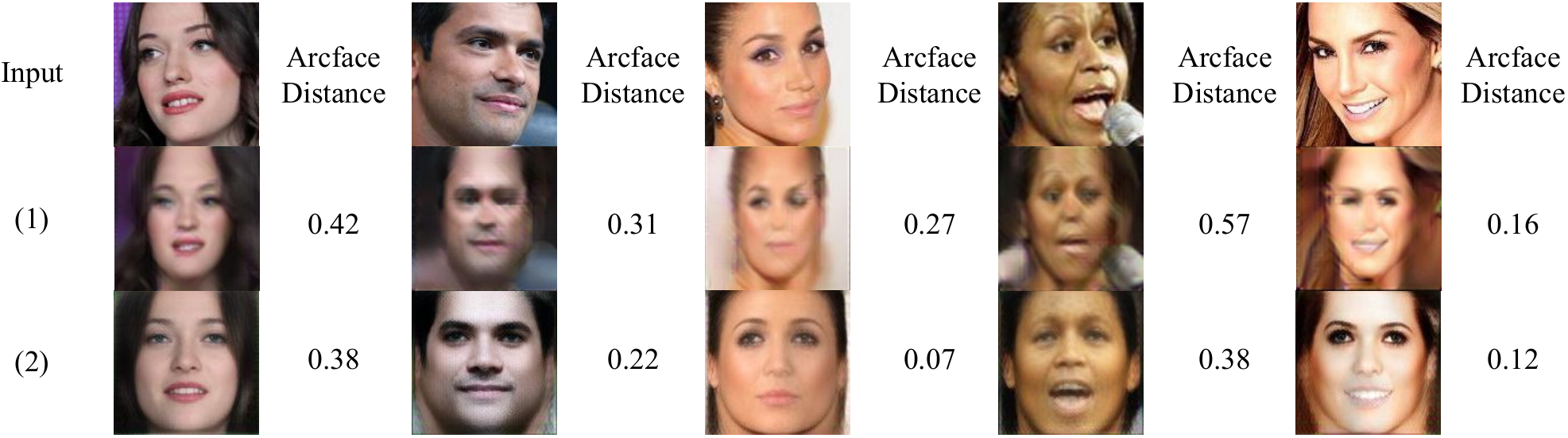}
	
	\caption{Face frontalization results. (1) is the output of the state-of-the-art method FaceCycle  \cite{Chang_2021_ICCV}. (2)  Our model can restore the complete front face with lower Arcface distance.}
	\label{fig:front}
\end{figure}

\begin{table*}[t]
	\centering  
 	\caption{Facial expression recognition on RAF-DB and AffectNet. Pose 30$^{\circ}$ and Pose 45$^{\circ}$ are two subsets of AffectNet with large head poses. We compare the classification accuracy(\%)/f1-score(\%) for all models. Bold text indicates the best results. * means the results are taken from the paper.}
	\begin{tabular}{@{}llcccccccc@{}}
		\toprule
		\multirow{2}{*}{Type} & \multirow{2}{*}{Method} & \multicolumn{2}{c}{RAF-DB} & \multicolumn{2}{c}{AffectNet} & \multicolumn{2}{c}{Pose 30$^{\circ}$} & \multicolumn{2}{c}{Pose 45$^{\circ}$}  \\ \cmidrule{3-10}
        & & Acc & F1 & Acc & F1 & Acc & F1 & Acc & F1 \\
        \midrule
		\multirow{7}{*}{Contrastive}&SimCLRv2 \cite{chen2020simple}  & 64.47&47.48 & 40.15&38.86	& 37.69&36.57	& 37.84&37.41	 \\
		&FAb-Net \cite{Wiles2018SelfsupervisedLO}   & 65.54&50.40  & 36.92&34.83 & 34.55&33.33 & 33.18&31.54  \\
		&TCAE \cite{Li2019SelfSupervisedRL}   & 69.52&57.90 & 38.77&37.14 & 36.56&35.24 & 35.60&33.67  \\
		&Temporal \cite{Lu2020SelfSupervisedLF}     & 56.94&42.92  & 34.20&32.53 & 34.78&33.47 & 34.50&32.25   \\
		&FaceCycle \cite{Chang_2021_ICCV}     & 71.51&60.80  & 41.04&39.70 & 38.50&37.10 & 36.48&34.90  \\
        &SSLFER \cite{Shu2022RevisitingSC}  & 55.37&40.37  & 35.90&34.49 &	34.72&33.69 & 34.06&32.15 \\ 
        &PCL* \cite{liu2023pose}  &74.47 & - & - & - & - & - & - & - \\
        &PCL \cite{liu2023pose}  &67.07&55.71 & 36.15&34.74 & 35.94&34.94& 34.72&33.58 \\
        \midrule
        \multirow{3}{*}{Generative}&MAE \cite{he2022masked}  & 62.61&48.93 & 37.40&33.99 & 35.78&33.05 & 35.05&32.06 \\
		&Unsup3D \cite{Wu2020UnsupervisedLO} & 69.42&55.50  & 41.07&39.26  & 38.89&38.14 &	36.70&35.25 \\
		&\textbf{Ours}   & \textbf{75.26}&\textbf{64.91 }  & \textbf{44.42}&\textbf{43.60 } & \textbf{41.95}&\textbf{41.69} & \textbf{40.43}&\textbf{39.45}   \\ \bottomrule
	\end{tabular}
	\label{tab:fer}
\end{table*}

\begin{table}[t]	
 \caption{Accuracy(\%) comparison on face verification. We show the mean and standard error for the results of 10-fold cross-validation.}
\centering
\begin{tabular}{@{}llcc@{}}
\toprule
Type& Method & LFW & SLLFW \\ \midrule
\multirow{7}{*}{Contrastive}&SimCLRv2  \cite{chen2020simple}         & 63.38$\pm$2.53	& 54.40$\pm$2.45          \\

&FAb-Net  \cite{Wiles2018SelfsupervisedLO}           & 69.57$\pm$1.54        & 56.62$\pm$2.13          \\
&TCAE  \cite{Li2019SelfSupervisedRL}           & 71.32$\pm$1.67        & 58.05$\pm$1.96          \\
&Temporal  \cite{Lu2020SelfSupervisedLF} & 65.95$\pm$2.89	& 57.03$\pm$3.14 \\
&FaceCycle  \cite{Chang_2021_ICCV}       &   71.20$\pm$1.90        & 59.60$\pm$2.73         \\
&SSLFER  \cite{Shu2022RevisitingSC}   & 62.38$\pm$2.88 & 54.55$\pm$2.33 \\
&PCL  \cite{liu2023pose}   & 67.50$\pm$2.31 & 58.12$\pm$0.89 \\ \midrule
\multirow{3}{*}{Generative}&MAE \cite{he2022masked}         & 62.60$\pm$2.75	& 53.05$\pm$2.25          \\ 
&Unsup3D  \cite{Wu2020UnsupervisedLO} & 71.37$\pm$1.99 & 58.70$\pm$1.62 \\ 
&\textbf{Ours}              & \textbf{72.05$\pm$2.57}        & \textbf{59.72$\pm$1.33}          \\ \bottomrule
\end{tabular}
\label{tab:veri}
\end{table}

\subsection{Evaluation of Interpretable Representations}

\cref{fig:disentangle} shows a qualitative comparison of the extracted face representations. The facial identity and expression of the face extracted by FaceCycle \cite{Chang_2021_ICCV} are not fully disentangled. In contrast, our method extracts face material and shape representations disentangled from pose and illumination. Furthermore, our model can disentangle facial expressions and identity better. Compared to FaceCycle's neutral face, our generated texture of facial identity is closer to the neutral state.

\cref{fig:front} shows the results of face frontalization compared with FaceCycle \cite{Chang_2021_ICCV} on the CelebA dataset. With our method, we can synthesize a frontal image of a human face with the extracted facial texture and depth by setting the pose in the canonical view. We compare the cosine distance of Arcface \cite{deng2018arcface} embedding between the frontalized face with the original face. The generated faces of FaceCycle \cite{Chang_2021_ICCV} are blurry and have artifacts. Our model can make light and pose disentangled from the environment and restore the color information of the occluded face, which achieves lower Arcface distances.

\subsection{Evaluation on Facial Expression Recognition} 

Facial expression recognition(FER) is a task that divides the expressions on a face image into various categories. We adopt the in-the-wild image facial expression datasets AffectNet \cite{Mollahosseini2019AffectNetAD} and RAF-DB \cite{li2017reliable,li2019reliable}. We adopt seven official facial expressions for RAF-DB and eight for AffectNet. We have cropped the images using provided face boxes for AffectNet and used aligned images for RAF-DB. Because of the unbalanced labels, we follow previous work \cite{li2022towards,xue2021transfer} to downsample the AffectNet dataset. For AffectNet, the test annotation is not released, and we test on the validation set. Our model uses the expression bias $\Delta_{exp}$ of texture and shape and the original $Z_{exp}$ for linear probing. And the linear layer is trained for 20 epochs for each dataset.

\cref{tab:fer} reports that our model achieves the best results on the listed datasets, with 75.26\% on RAF-DB, and 44.42\% on AffectNet. Also, our method has a 3.75\% advantage on RAF-DB and 3.35\% on AffectNet compared to the state-of-the-art method. Our model achieves state-of-the-art performance compared with other unsupervised methods. We notice that PCL achieves a result far from the original paper, where the linear probing has a long training time of 300 epochs. Furthermore, we follow the paper \cite{Wang2020RegionAN} to test facial pose robustness on two subsets of the AffectNet with head pose at 30 and 45 degrees separately. Since we disentangle the pose, our model performs excellently, with 41.95\% accuracy on 30 degrees and 40.43\% on 45 degrees. 

\subsection{Evaluation on Face Verification} 
Face verification is the task of comparing a candidate's face with another face and verifying whether it matches. We use LFW \cite{LFWTech,LFWTechUpdate} and SLLFW \cite{deng2017fine,Zhang2016Fine} as the test datasets. The datasets have 6000 pairs of face images separately. Different from LFW, SLLFW contains similar-looking face image pairs by human crowdsourcing. We crop images using provided landmarks  \cite{wang2021facex}. 

We follow the 10-fold cross-validation in the papers \cite{LFWTech,LFWTechUpdate}. Each image's facial embedding is extracted first with each model. Then we divide the dataset into ten parts with 300 positive and 300 negative pairs each. We use nine of them to train the linear model and one to test. We repeat this process for cross-validation and report the averaged results in ten folds. Our model uses identity latent $Z_{id}$ of texture and shape and the original $Z_{exp}$ for linear probing. 

\cref{tab:veri} shows the evaluation result on the face verification task. Our model disentangles the identity from the expression and considers facial shape and texture separately to get better discrimination. Moreover, the representations are not distracted from expression and external factors, such as light and pose. Therefore, we achieve the best performance among unsupervised methods with an accuracy of 72.05\% on LFW and 59.72\% on SLLFW.

\begin{table}[t]
\caption{Ablation study on pixel space disentangling. We show the classification accuracy/f1-score. }
	\centering
	\begin{tabular}{@{}lcc@{}}
		\toprule
	Method	 & RAF-DB & AffectNet \\ \midrule
w/o Light	&64.21/50.67	&33.05/30.43 \\
w/o Pose	&67.07/55.02	&37.42/35.65 \\
w/o Shape	&68.38/58.22	&37.67/36.37 \\
w/o Texture	&72.58/60.73	&41.22/39.24 \\
\textbf{Full Representations}	    &\textbf{75.26/64.91 }   & \textbf{44.42/43.60 }  \\ \bottomrule
	\end{tabular}
	\label{tab:ablation}
\end{table}

\begin{figure}[t]
	\centering
	\includegraphics[width=\linewidth]{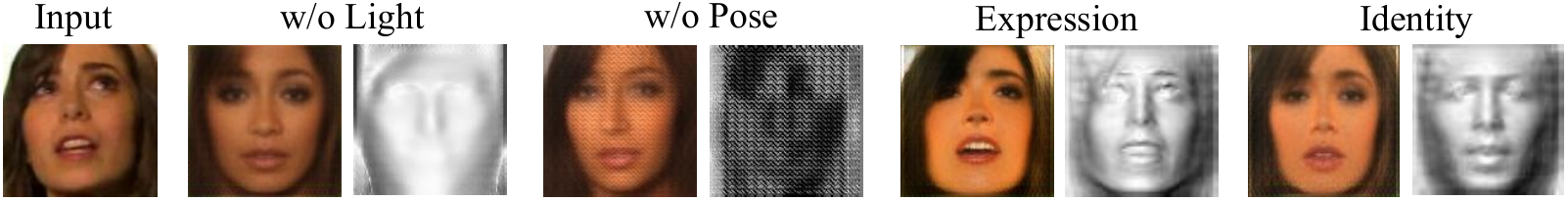}
	\caption{Ablation study of pose, light, expression and identity.}
	\label{fig:ablation}
\end{figure}

\begin{table}[t]
\caption{Ablation study on latent space disentangling. We show the classification accuracy/f1-score for RAF-DB and AffectNet, and accuracy score for LFW and SLLFW.}
	\centering
	\begin{tabular}{@{}lcccc@{}}
		\toprule
	Expression	 & RAF-DB & AffectNet & LFW & SLLFW\\  \midrule
	  None         	&69.42/55.50	&41.07/39.26 & 71.37$\pm$1.99 & 58.70$\pm$1.62  \\
		Encoder-based method                 & 73.04/61.37   & 41.94/40.70  & 69.00$\pm$2.56   & 57.18$\pm$1.27       \\
        RDM on Texture & 73.98/63.89  & 42.80/41.82	& \textbf{72.27$\pm$2.69}	& 59.37$\pm$1.83\\
		RDM on Shape     & 74.57/64.13   &  43.47/42.54  & 71.55$\pm$2.31&	58.50$\pm$1.36\\
		\textbf{RDM on T \& S}  &\textbf{75.26/64.91 }   & \textbf{44.42/43.60 } & 72.05$\pm$2.57        & \textbf{59.72$\pm$1.33} \\  \bottomrule
	\end{tabular}
	\label{tab:ablation_latent}
\end{table}

\section{Ablation Study}
\label{sec:ablation}

\subsection{3D Latent Autoencoding}

We separately remove each 3D face representation from the full model for the ablation study. We remove the shape and texture encoder in the trained model. However, light and pose affect the disentangling of facial shape and texture in the training process. Thus, we remove the light and pose encoders before training. \cref{tab:ablation} report the results of facial expression recognition on RAF-DB \cite{li2017reliable,li2019reliable} and AffectNet \cite{Mollahosseini2019AffectNetAD}.  The effect of discarding shape and texture latent is 2.68\% and 6.88\% separately for RAF-DB. The facial shape is more crucial for facial expression recognition. Moreover,  \cref{fig:ablation} shows the effect of removing pose and light. As the light and pose of the face image change greatly, the model converges badly without these two representations. 

\subsection{Latent Space Disentangling}

\begin{figure}[t]
  \centering
  \includegraphics[width=0.7\linewidth]{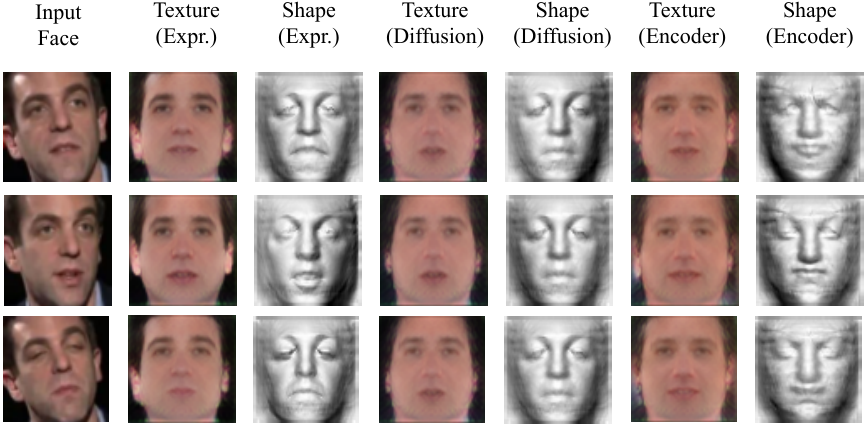}
   \caption{Ablation study of the diffusion model. The encoder-based model have worse ability to generate the identity.}
   \label{fig:predict}
\end{figure}

We also explore an intuitive alternative for latent diffusion. A ResNet  \cite{He2016DeepRL} encoder can directly predict the facial identity latent. We experiment with the two methods with the same first-stage model. We show the comparison of generated face identity on VoxCeleb \cite{Nagrani2017VoxCelebAL} dataset test set in \cref{fig:predict}. Since the frames are randomly sampled in training, the encoder-based model has a noisy target and generate more artifacts. In contrast, the diffusion-based method is trained to generate from a distribution and has better fidelity. 

\cref{tab:ablation_latent} compare different latent disentangling methods for the facial expression classification on RAF-DB and AffectNet and facial verification on LFW and SLLFW. The encoder-based method can also disentangle the representation and improve the performance. However, with the diffusion model, the performance boost of latent disentangling is more significant than the encoder-based method, with a 2.24\% accuracy advantage on RAF-DB and 2.48\% on LFW. Furthermore, we experimented with different diffusion models for different tasks. For facial expression recognition, the disentangling for shape is more important. In contrast, the disentangling for texture has a better improvement on face verification.

\begin{table}[t]
\centering
\caption{Disentangling capabilities comparison. The embedding size is counted for downstream tasks. Our method disentangles more facial factors than previous works while remaining reasonable embedding size. }
\begin{tabular}{@{}lcccccc|c@{}}
\toprule
Method &  Identity  & Expression & Light & Pose & Texture & Shape & Embedding Size\\ \midrule
FAb-Net \cite{Wiles2018SelfsupervisedLO}    & \checkmark & \checkmark & & \checkmark &  & & 256\\
TCAE \cite{Li2019SelfSupervisedRL}     & & \checkmark & & \checkmark &  & &256\\
Temporal \cite{Lu2020SelfSupervisedLF}    & & \checkmark & & &  & &256  \\
FaceCycle \cite{Chang_2021_ICCV}     & \checkmark & \checkmark & & \checkmark&  & & 6272\\
SSLFER \cite{Shu2022RevisitingSC} &  & \checkmark& &  &  & & 128\\
PCL \cite{Shu2022RevisitingSC} & \checkmark & \checkmark& & \checkmark &  & & 2048\\
Unsup3D \cite{Wu2020UnsupervisedLO}    &  &  & \checkmark & \checkmark & \checkmark &  \checkmark & 512 \\
\textbf{Ours}     & \checkmark & \checkmark & \checkmark & \checkmark & \checkmark & \checkmark & 1024\\ \bottomrule
\end{tabular}
  \label{tab:related}
\end{table}

\section{Related Work}
\label{sec:related}

\paragraph{Facial Representation Learning} Facial representation learning aims to get better representations of human faces and facilitate downstream tasks, which is usually performed in unsupervised manner. Recent works leverage the image pretraining methods like contrastive learning  \cite{caron2020unsupervised,He2020MomentumCF} for an implicit representation from various datasets. FAb-Net \cite{Wiles2018SelfsupervisedLO} learns 2D face motion heat maps from video to learn face representation, while TCAE \cite{Li2019SelfSupervisedRL} divides facial movements into action unit (AU)-related and pose-related ones to disentangle facial representations. Lu et al.  \cite{Lu2020SelfSupervisedLF} proposed a model to learn facial features using video contrastive learning algorithms, while FaceCycle \cite{Chang_2021_ICCV} progressively disentangle face representations by mining the cyclic consistency of expression and identity. SSLFER  \cite{Shu2022RevisitingSC} and PCL \cite{liu2023pose} further disentangles facial expressions and poses using a contrastive learning schema. In contrast, our model adopts a generative fashion of representation learning, which is more interpretable and achieve better performance. We also compare the disentangling capability in \cref{tab:related}.

\paragraph{3D Face Modeling} 3D face modeling aims to optimize the alignment of face images. One of the most widely used methods is the 3D Morphable Model (3DMM)  \cite{Blanz1999AMM}, which models the human face as a linear parameterized 3D model. Many improvements  \cite{Feng2021LearningAA, deng2019accurate, Tewari2018SelfSupervisedMF, Tran2021OnL3, tewari2019fml} have been made to 3DMM for improved alignment accuracy and natural expression. However, this method requires extensive manual annotation from face scanning, which is expensive and laborious. On the other hand, unsupervised face models, such as Unsup3D  \cite{Wu2020UnsupervisedLO} and Lifting Autoencoders  \cite{Sahasrabudhe2019LiftingAU}, have recently gained attention because they do not require face annotations. However, these models have limited design for self-supervised facial representation and has a subpar performance with direct use.

\paragraph{Latent Diffusion Models} Diffusion models \cite{Ho2020DenoisingDP,SohlDickstein2015DeepUL} have emerged as a popular alternative to GANs for image synthesis \cite{dhariwal2021diffusion,Nichol2021GLIDETP} in the pixel space. These models have also shown successful results in various domains, such as video generation \cite{Wu2022TuneAVideoOT,Ho2022VideoDM}, image restoration \cite{Kawar2022DenoisingDR,luo2023refusion}, semantic segmentation \cite{Xu2023OpenVocabularyPS,Baranchuk2021LabelEfficientSS}, and natural language processing \cite{li2022diffusion}. In the diffusion-based framework \cite{Ho2020DenoisingDP}, models are trained on images using score-matching objectives at different noise levels, and sampling is achieved through iterative denoising.  \cite{Rombach2021HighResolutionIS} have demonstrated the potential of learning the diffusion model in latent space defined by a pretrained autoencoder \cite{esser2021taming}, achieving better quality than ever. Inspired by the success, we explore the use of latent diffusion models for facial representation learning.

\section{Conclusion and Discussion of Broader Impact}

In this paper, we propose LatentFace, a novel generative framework for self-supervised facial representations. We suggest that the disentangling problem can be also formulated as generation objectives and propose the solution using a 3D-aware latent diffusion model. Experimental results demonstrate that our approach can achieve SOTA performance with high interpretability.

Although our method has achieved superior performance in self-supervised facial representation learning, interpreting a face with a large deflection angle is still not easy due to occlusion. Our model can disentangle the face with up to 60-degree deflection and will support larger angle with proper data. Meanwhile, the extracted representations will reflect the real facial shape and texture, which may be used for further face edition and generation, causing potential application risks. For real application, the decoders can be removed in concerns of privacy.

\bibliographystyle{plain}
\bibliography{sample-base}

\appendix

\section{Facial Expression Analysis}

\begin{figure}[h]
	\centering
	\includegraphics[width=0.6\linewidth]{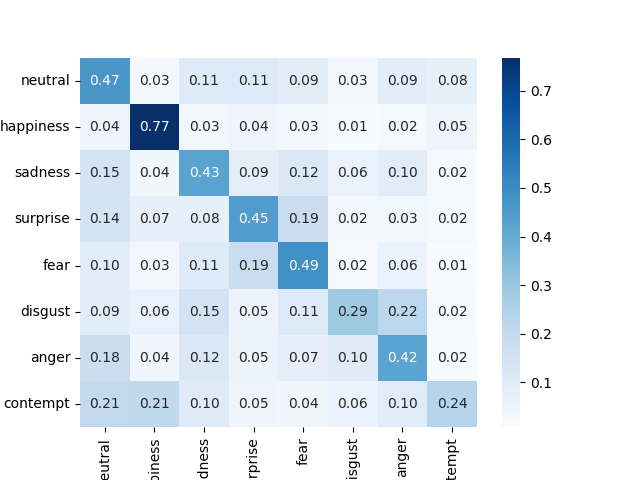}
	
	\caption{Confusion matrix analysis on AffectNet dataset. We show actual classes in y-axis and predicted classes in x-axis.}
	\label{fig:affect}
\end{figure}

\begin{figure}[h]
	\centering
	\includegraphics[width=0.6\linewidth]{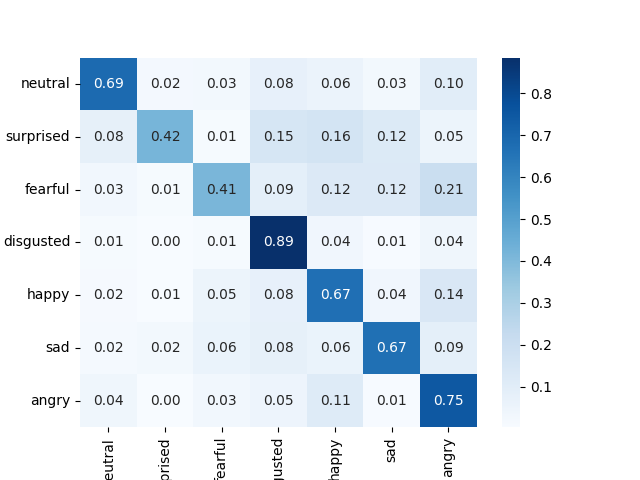}
	
	\caption{Confusion matrix analysis on RAF-DB dataset. We show actual classes in y-axis and predicted classes in x-axis.}
	\label{fig:rafdb}
\end{figure}

For facial expression recognition, we further evaluate the confusion matrix of our predictions, as depicted in \cref{fig:affect} and \cref{fig:rafdb}. In AffectNet, the classes with the lowest accuracy are 'disgust' (29\%) and 'contempt' (24\%). In RAF-DB, the classes with the lowest accuracy are 'surprised' (42\%) and fearful (41\%). We hypothesize that the classification performance is impacted by the dataset's imbalance. Notably, the 'disgust' and 'contempt' classes in AffectNet comprise the smallest proportion of the dataset, each representing only 5\% of the samples in the 'neutral' category. We propose that balancing the downstream dataset might improve classification accuracy.

\section{Network Details}

\begin{figure}[t]
	\centering
	\includegraphics[width=0.6\linewidth]{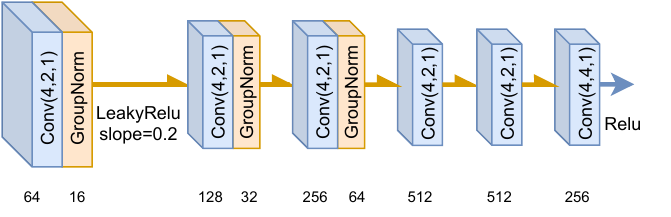}
	
	\caption{Feature Encoder Architecture. $Conv(a,b,c)$ indicates that the kernel size of the convolutional layer is $a$, the stride is $b$, and the padding is $c$. The number below the convolutional layer indicates the number of convolution kernels. The number below the group normalization layer indicates the number of groups. The yellow arrow indicates Leaky Relu, with a slope of 0.2. And the blue arrow is Relu.}
	\label{fig:feature_encoder}
\end{figure}

\begin{figure*}[t]
	\centering
	\includegraphics[width=\linewidth]{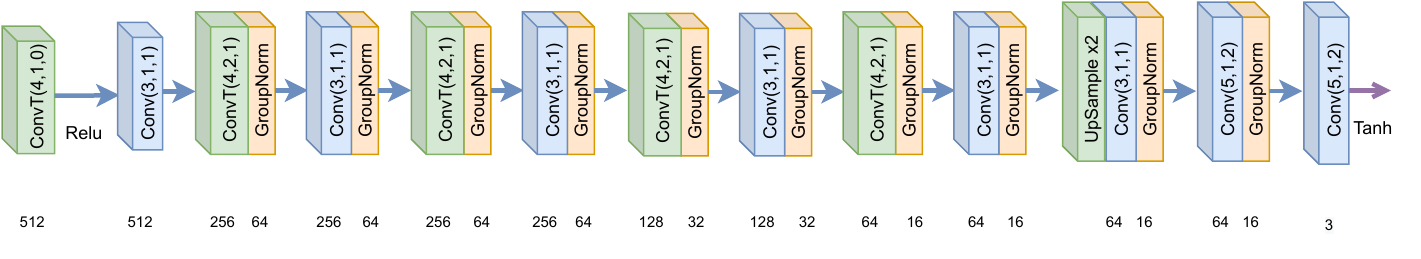}
	
	\caption{Decoder Architecture. $Conv(a,b,c)$/$ConvT(a,b,c)$ indicates that the kernel size of the convolutional layer/transposed convolutional layer is $a$, the stride is $b$, and the padding is $c$. The number below the module indicates the number of convolution kernels. And The number below the group normalization layer indicates the number of groups. }
	\label{fig:decoder}
\end{figure*}

For the architecture of the encoders, we partition them into feature encoders and numerical encoders by structure. The feature encoder refers to the facial texture encoder and the facial shape encoder. They work with the decoder to encode the two-dimensional features of the face.  Figure~\ref{fig:feature_encoder} shows the detailed structure. The two encoders have the same architecture, and what they produce is a 256-dimensional vector. We use a fully convolutional architecture to reduce the influence of our features on the position of the pose. Our decoder uses stacked convolutional layers, transposed convolutional layers, and group normalization layers. Figure~\ref{fig:decoder} shows the detailed structure. The network uses a 256-dimensional vector as input. The facial texture decoder produces a 3-channel two-dimensional image output, while the facial shape decoder produces 1 channel. The final texture map and depth map will be scaled by Tanh to the range of -1 to 1.

\begin{figure}[t]
	\centering
	\includegraphics[width=0.6\linewidth]{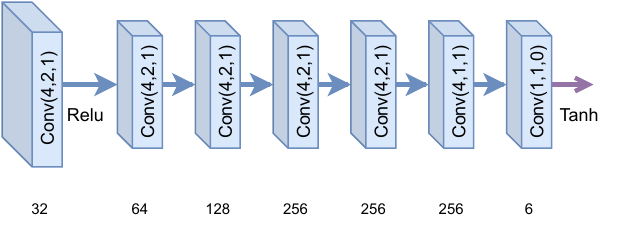}
	
	\caption{Numerical Encoder Architecture. It is a fully convolutional network. $Conv(a,b,c)$ indicates that the kernel size of the convolutional layer is $a$, the stride is $b$, and the padding is $c$. The number below the module indicates the number of convolution kernels. The blue arrow is Relu. And the purple arrow is Tanh.}
	\label{fig:numerical_encoder}
\end{figure}

\begin{figure*}[t]
	\centering
	\includegraphics[width=\linewidth]{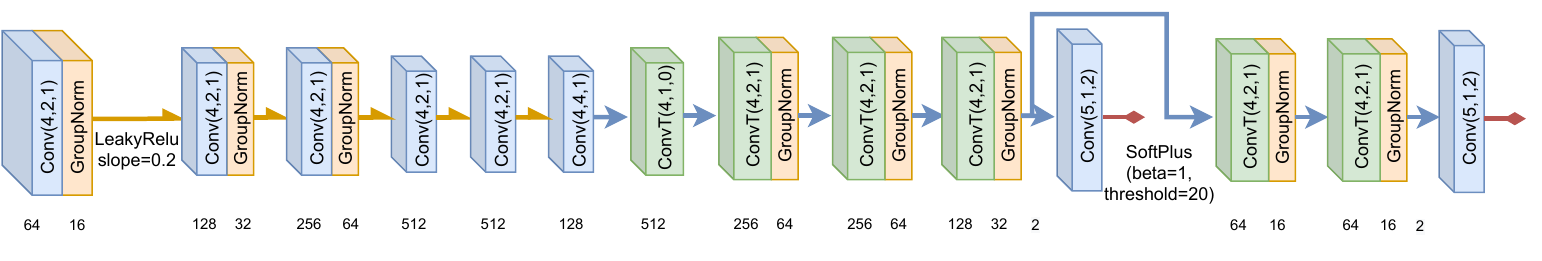}
	
	\caption{Confidence Map decoder Architecture. The yellow arrow indicates Leaky Relu, with a slope of 0.2. And the blue arrow is Relu. The red arrow is a SoftPlus operator with beta=1 and threshold=20. The shorter path is for the feature level loss and the longer path is for the photometric loss.}
	\label{fig:conf}
\end{figure*}

\begin{figure*}[t]
	\centering
	\includegraphics[width=\linewidth]{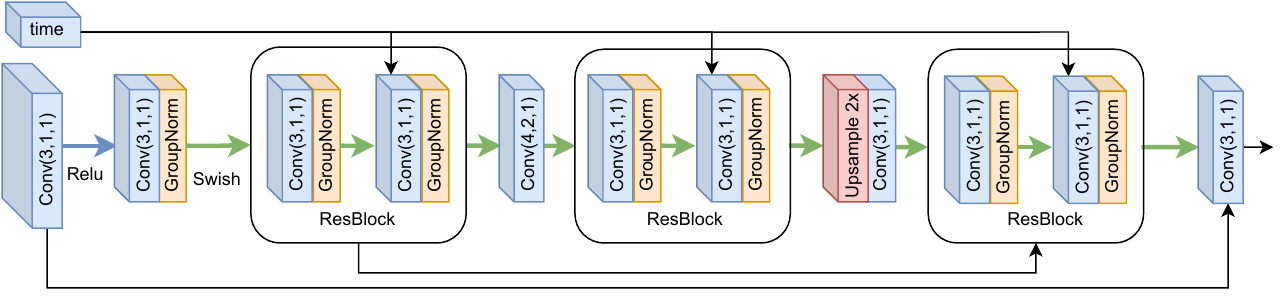}
	
	\caption{Denoising U-Net Architecture. $Conv(a,b,c)$ indicates that the kernel size of the convolutional layer is $a$, the stride is $b$, and the padding is $c$. The number of convolution kernels are set to 512, 128 and 256. The blue arrow is Relu while the green arrow is Swish Gate. We use sinusoidal timestep embeddings defined in \cite{Ho2020DenoisingDP}. We omit the skip connections in the ResBlocks\cite{He2016DeepRL} for a clear view.}
	\label{fig:conf}
\end{figure*}

The numerical encoder refers to the light and head pose encoder.  Figure~\ref{fig:numerical_encoder} shows the detailed structure. They generate corresponding values separately to set the rendering condition.  These two encoders have the same structure, except that the output vector dimensions are slightly different. The head pose encoder has 6 dimensions, which are the translation vector of x, y, and z and the rotation angle of yaw, roll, and pitch. The light encoder has 4 dimensions: ambient light parameters, diffuse reflection parameters, and two light directions $(x, y)$. We will first use Tanh to scale the final outputs to $[-1, 1]$, and then map them to the corresponding space.

Following \cite{Wu2020UnsupervisedLO}, we use a similar structure as the depth encoders and decoders to generate the confidence map for the self-calibration of the loss function. Figure~\ref{fig:conf} shows the network structure. We use the \textit{relu3\_3} feature extracted by a pretrained VGG\cite{Simonyan2015VeryDC} network for feature-level loss. The VGG networks can also be replaced with self-supervised discriminators as demonstrated by \cite{Wu2020UnsupervisedLO}.

For the architecture of the denoising network, we use one 3 blocks with two ResNet\cite{He2016DeepRL} layers in each block. The number of convolution kernels are set to for downsample, middle and upsample. Figure~\ref{fig:conf} shows the network structure. The input 256-dimensional noise is concatenated with the facial texture latent or the facial shape latent.  Therefore, the input size is 512.  The output size is 256 dimensional, the same as the input noise. The facial texture latent diffusion model and facial shape latent diffusion model are learned separately.

\section{Loss Function}
\label{sec:loss}

In the first stage, our reconstruction loss is composed of a pixel-level loss $L_{p}(\hat{I}, I)$, a feature-level loss $L_{f}(\hat{I}, I)$, which measure the photometric discrepancy in pixel level, and feature level separately, which is also used by many works\cite{deng2019accurate,Feng2021LearningAA, Wu2020UnsupervisedLO}.
\begin{align}
	L_{p}(\hat{I},I) &= L_{\mathit{conf}}(\hat{I},I,\sigma_p) \\
	L_{f}(\hat{I},I) &= L_{\mathit{conf}}(\mathit{conv}(\hat{I}),\mathit{conv}(I),\sigma_f) ,
\end{align}
where $I$ represents the input image and $\hat{I}$ is the reconstructed image. The reconstruction losses are constrained with estimated confidence maps $\sigma$ to make the model self-calibrate \cite{Kendall2017WhatUD}. 
\begin{equation}
	\label{eqa:conf}
	L_{\mathit{conf}}(\hat{I},I,\sigma)=-\frac{1}{|\Omega|}\sum_{u,v\in \Omega}\ln \frac{1}{\sqrt2\sigma_{u,v}}
	\exp -\frac{\sqrt2 (\hat{I}-I)_{u,v}}{\sigma_{u,v}}
\end{equation}
where $L$ is the reconstruction loss, and $u,v$ is the pixel on the rendered face mask $\Omega$. We separately predict the confidence map $\sigma_p$ and $\sigma_f$ from the image with an encoder-decoder structure for the pixel-level and feature-level loss. A pretrained VGG\cite{Simonyan2015VeryDC} is leveraged for the low-level feature extraction network $\mathit{conv}$. 

The reconstruction loss $L(\hat{I}, I)$ can be expressed as a linear combination of the pixel-level loss, and the feature-level loss.
\begin{align}
L(\hat{I},I) = L_{p}(\hat{I},I) + \lambda_{f} L_{f}(\hat{I},I) + \lambda_{flip}( L_{p}(\hat{I}',I) + \lambda_{f} L_{f}(\hat{I}',I)) 
\end{align}
where $\lambda_{f}$ are the weight for the feature-level. Moreover, we also calculate the reconstructed image $\hat{I}'$ of the horizontally flipped shape and texture code for the symmetric loss function. $\lambda_{flip}$ is the weight for flipped reconstruction loss. The hyperparameters in the loss function are set as follows: $\lambda_f = 1, \lambda_{flip}=0.5$.

For the second stage, we optimize the denoising U-Net $\mathcal{E}_{r}$ with \cref{eqa:unet}. Following \cite{Ho2022VideoDM, Rombach2021HighResolutionIS}, we use the SNR weighting strategy, which uses different weights for different steps. For given target latent $Z_{0}$ and our predictions $\hat{Z}_0$, the denoising loss can be expressed as:
\begin{equation}
    L(\hat{Z}_{0},Z_{0}) = E_{Z_0,\epsilon,t}\left[w_t\left\|Z_0-\mathcal{E}_{r}\left(Z_t, t, Z_{exp}\right)\right\|^{2}\right]
\end{equation}
where $w_t$ is the SNR weight depended on step $t$, which can be expressed as $w_t = \frac{\alpha_t^2}{1-\alpha_t^2}$. $\alpha_t$ is the noise schedule in \cite{Ho2020DenoisingDP}. Like the latent diffusion models\cite{Rombach2021HighResolutionIS}, the encoders and decoders are frozen when training the denoising U-Net.

\section{Rendering Pipeline}

In order to render a face image, we need to know: facial shape, facial texture, lighting, and head pose. The facial shape is represented as a two-dimensional single-channel matrix, which represents the depth map of the face. We define a grid, the size of the grid is the same as the image, which is 64 $\times$ 64. Their x-axis and y-axis coordinates are scaled to between -1 and 1, and then the z-axis coordinates come from the depth map. In this way, we get a three-dimensional model of the human face and the normal of every point for later calculation.The facial texture is represented as a 3-channel 2D matrix in the renderer. It indicates the diffuse reflectance of each face grid point of RGB rays.

Lighting includes ambient light intensity, diffuse reflection intensity and light direction x, y. We are modeling directional light, so only two variables are needed to describe the direction of light. According to the illumination equation \cite{Phong1975IlluminationFC}, we already know the ambient light intensity $k_{\text{a}}$ , diffuse reflection intensity $k_{\text{d}}$, light direction vector ${\mathbf {L}}_{m}$, and the surface normal vector ${\mathbf {N}_p}$ of each point, and the diffuse reflection coefficient $k_{\text{d},\text{p}}$. Our model ignores the specular reflection of the human face. In most cases, the specular reflection coefficient of the human face is small enough to omit. 

The head pose is actually described as the camera parameters. The movement of the head relative to the camera and the change of the camera's shooting angle are mutual processes. Thus, we only need to control the camera's shooting displacement vector under x, y, z, and the rotation angle yaw, roll, pitch. We can apply these parameters to the camera imaging equation as below. The displacement vector is $\mathbf{t}$ and the rotation vector is $\mathbf{R}$. And we define the camera matrix $\mathbf{K}$ as $diag(f,f,1)$, where $f$ is the focal length. The focal length can be calculated as $\frac1f = 2\tan\frac{\theta_{FOV}}{2}$, where $\theta_{FOV}$ is set as 10 degrees.

In short, we build a three-dimensional skeleton of the face shape at first, and then map the face material to the three-dimensional skeleton. Then we use the light information to determine the color of the face. Finally, we use the camera formula to get the image we took at a specific angle.

\end{document}